\documentclass{article}

\usepackage[preprint]{corl_2026}
\usepackage{marvosym}
\usepackage[noblocks]{authblk}

\title{Trajectory-Level Automatic Curriculum Learning for Legged Locomotion on Unstructured Terrain}

\renewcommand{\Affilfont}{\normalfont\scriptsize}

\makeatletter
\renewcommand{\AB@affilsepx}{\protect\quad\protect\Affilfont}
\makeatother

\author[1]{Rocky Liu}
\author[1]{Tengyu Liu}
\author[1]{Baoxiong Jia}
\author[4]{Fangwei Zhong}
\author[1,2]{Xinyi Tong}
\author[3]{Hongzhao Xie}
\author[1]{Siyuan Huang\,\Letter}

\affil[1]{National Key Laboratory of General Artificial Intelligence, BIGAI}
\affil[2]{Communication University of China}
\affil[3]{Shanghai Jiao Tong University}
\affil[4]{Beijing Normal University}

\hypersetup{pdfauthor={Rocky Liu, Tengyu Liu, Baoxiong Jia, Fangwei Zhong, Xinyi Tong, Hongzhao Xie, Siyuan Huang}}

\usepackage{amssymb}
\usepackage{amsmath}
\usepackage{graphicx}
\usepackage{booktabs}
\usepackage{subcaption}
\usepackage{xcolor}
\usepackage{enumitem}
\usepackage{wrapfig}   
\usepackage{booktabs}

\definecolor{randst_color}{RGB}{216,94,26}
\definecolor{ours_color}{RGB}{38,141,69}
\definecolor{hcle_color}{RGB}{109, 87, 138}

\newcommand{\ours}{\textcolor{ours_color}{TACL}}
\newcommand{\randst}{\textcolor{randst_color}{RandST}}
\newcommand{\hcle}{\textcolor{hcle_color}{HCL-E}}
\newcommand{\ourname}{TACL}
\begin{document}
\maketitle

\begin{center}
    \centering
    \vspace{-25pt}
    \includegraphics[width=\textwidth]{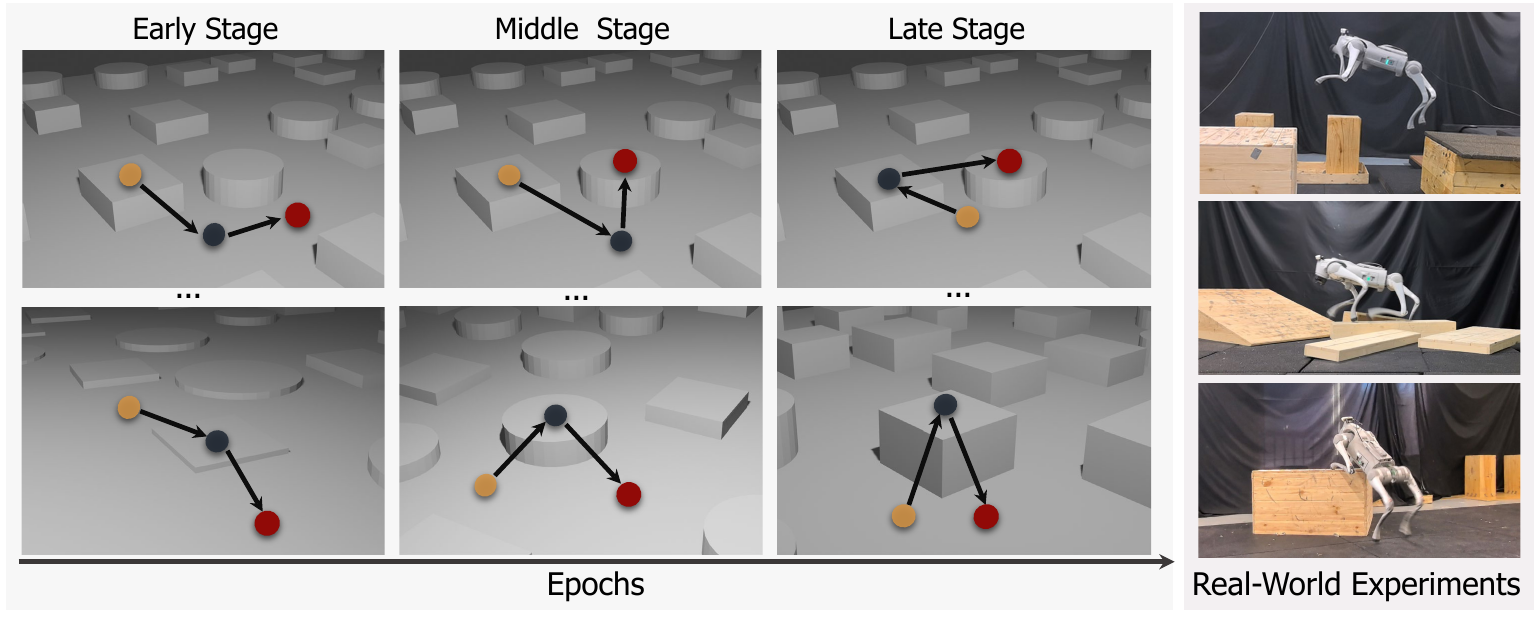}
    \captionof{figure}{Our method generates effective curricula on unstructured terrain. As training progresses, the sampled tasks evolve with the policy capability, moving toward taller platforms, harder approach angles, descents from elevated structures, and gap-crossing between high platforms. Experiments show that our method improves success rates across terrain tasks, learns more robust behaviors such as platform climbing from diverse approach directions, and supports sim-to-real transfer.
}
    \label{fig:teaser}
\end{center}

\begin{abstract} 
Training locomotion policies for complex unstructured terrain requires a curriculum to avoid early exploration failures. However, since unstructured terrain lacks explicit difficulty ordering for curriculum design, existing methods resort to heuristic curricula over parameterized terrains. This abstraction limits generalization, as policies can overadapt to near-fixed perceptual patterns. To address this, we propose \textbf{\ourname{}}, an \textbf{T}rajectory-level \textbf{A}utomatic \textbf{C}urriculum \textbf{L}earning framework that generates training tasks directly from unstructured terrain maps. At each curriculum update, the evaluator learns a difficulty function for the current policy that maps a given trajectory task to a difficulty score. The sampler then proposes new trajectories guided by the learned evaluator as the curriculum for the next policy update. This forms a closed loop in which the curriculum is iteratively matched to the evolving policy. Quantitative and qualitative experiments show that \ourname{} continuously provides effective curricula on unstructured terrain, improving trajectory success rate by \(56.3\%\) over direct training without curriculum. Compared with handcrafted curriculum learning, our method improves success rate by \(18.5\%\) on the hardest terrain tasks and by up to \(39.74\%\) when evaluating traversal from diverse approach directions on the same obstacle type.

\end{abstract}

\keywords{Robot Locomotion, Curriculum Learning} 

\section{Introduction}
Learning robust locomotion policies for complex terrain with deep reinforcement learning (DRL) \citep{sutton1998reinforcement}. Among the success, curriculum learning is a key role to avoid conservative policies caused by early-stage exploration failures. However, unstructured maps do not provide an explicit criterion for judging which tasks are suitable for the current policy or how task difficulty should progress. Existing methods address this by imposing structured curriculum spaces.

Handcrafted curriculum learning (HCL) abstracts complex terrain into several parameterized geometric templates \citep{hoeller2024anymal, cheng2024extreme, wang2025beamdojo, luo2024pie, dong2025marg, makoviychuk2021isaac}, e.g., gap terrains, instantiates them along heuristic difficulty progressions, and pairs them with manually specified trajectories for policy training. Despite empirical success in structured settings, HCL requires substantial manual effort and produces limited task distributions, making policies overly adapted to the nearly fixed perceptual patterns along predefined trajectories. Automatic curriculum learning (ACL) reduces the need to manually specify difficulty progressions by adapting task sampling using policy performance or learning signals  \citep{li2026scaling, lee2020learning, ALLSTEPs, kim2025high}, but it still generate curricula within the same parameterized template spaces rather than directly on unstructured maps. In both HCL and ACL, this abstraction causes terrain information loss and narrows the task distribution, leaving the policy underexposed to the diverse trajectory tasks encountered in unstructured environments. To avoid these changalles that construct curricula directly on unstructured maps, the missing criterion must be learned at the trajectory level. Unlike HCL, which heuristically defines difficulty through predefined terrain-parameter levels, trajectory difficulty on an unstructured map depends on the terrain variations encountered along the path, the transition context, and the current policy capability. The key challenge is therefore to evaluate trajectory difficulty under current policy and sample learnable trajectory-level tasks on a map, without manually designing a terrain parameterization.

In this paper, we propose \ourname{}, an automated curriculum learning framework capable of generating trajectory tracking curricula that match the evolving competence of the policy, enabling effective learning directly on raw unstructured maps. The framework comprises two primary modules: the evaluator and the sampler. At each curriculum update, we roll out the current policy on trajectories randomly sampled from the map, collect success and failure outcomes, and represent each traversed segment by its transition context. We use the conditional variational autoencoder (CVAE) \citep{sohn2015learning} to encode these high-dimensional transition contexts into a compact latent space. The evaluator then learns a policy-conditioned difficulty function that predicts the difficulty of each trajectory from its encoded transition context. Guided by the learned evaluator, the sampler uses Metropolis-Hastings Markov Chain Monte Carlo (MH-MCMC) \citep{metropolis1953equation, hastings1970monte} to search for new waypoint trajectories whose predicted difficulty falls within a suitable range for the current policy. The generated trajectories are used as the curriculum for the next policy update, forming a closed loop in which tasks become progressively harder as the policy improves.

We evaluate our method on both an unstructured terrain map and the handcrafted curriculum map from Extreme Parkour \citep{cheng2024extreme}, comparing against random trajectory sampling and the HCL baseline. Quantitative and qualitative results show that our method generates effective curricula directly on unstructured terrain, achieving more roboust terrain traversal performance to HCL while improving trajectory-level generalization by \(39.5\%\). Overall, our contributions can be summarized as follows:
\begin{itemize}[leftmargin=*,noitemsep,nolistsep]
\item We propose \ourname{}, a closed-loop framework for robot locomotion policy that automatically generates curricula directly on unstructured terrain maps, removing the need for handcrafted terrain curricula and manually parameterized terrain templates.
\item We design a context-aware transition encoding and evaluator that accurately predicts task difficulty for current policy, providing reliable guidance for sampling capability-matched training tasks.
\item Extensive experiments demonstrate that our framework generates effective curricula directly on unstructured terrain maps, substantially improving policy capability over non-curriculum training and achieving more robust traversal than HCL across diverse approach directions.
\end{itemize}

\section{Related Works}
\paragraph{Handcrafted Curriculum Learning in Robot Locomotion.}
Handcrafted curriculum learning reduces the mismatch between task difficulty and policy capability by manually scheduling training tasks from easy to hard~\citep{wang2021survey}. In standard locomotion, this is often done by expanding command ranges, such as target velocity, heading, or motion amplitude, enabling skills such as basic locomotion~\citep{tidd2020guided, qin2019sim, kumar2021rma}, high-speed trotting~\citep{margolis2024rapid, he2024agile}, and jumping~\citep{atanassov2024curriculum}. For complex terrain traversal, prior works typically handcraft both terrain curricula and trajectory tasks. They parameterize obstacle families such as stairs, gaps, hurdles, and ramps, and advance the policy through manually specified difficulty sequences using heuristic performance thresholds~\citep{tidd2020guided, hoeller2024anymal, cheng2024extreme, wang2025beamdojo, he2025attention, rudin2022learning, luo2024pie, dong2025marg}. Other parkour-oriented methods rely on handcrafted constraints, terrain-specific experts, or predefined waypoint trajectories to learn maneuvers such as crossing gaps, climbing platforms, or traversing ramps~\citep{zhuang2023robot, zhuang2024humanoid}. While effective within designed terrain families, these curricula require human expertise to define terrain categories, difficulty progressions, success thresholds, and often the training trajectories themselves. The resulting task distributions can be limited or nearly fixed, making policies overly adapted to specific waypoint paths and perceptual patterns, which reduces transfer to regenerated or less structured terrain instances.

\paragraph{Automatic Curriculum Learning in Robot Locomotion.}
Automatic curriculum learning aims to reduce manual curriculum design by adaptively selecting training tasks according to the policy's learning state. A central challenge is to identify tasks that are neither too easy nor too difficult for the current policy. Existing approaches use reward prediction, temporal-difference (TD) error, regret, surprise, or learning progress as task-selection signals. HACL learns a reward predictor from historical data for task sampling~\citep{mishra2025hacl}, while unsupervised environment design methods generate increasingly difficult environments through adversarial or evolutionary processes~\citep{wang2019paired, parker2022evolving, dennis2020emergent, wang2025gacl}. TD-error- or surprise-based methods prioritize tasks with high prediction error~\citep{jiang2021prioritized, jiang2021replay, xu2025vertiselector}, but such signals can mix learnable uncertainty with irreducible noise and trap the policy in unproductive regions~\citep{portelas2020automatic, saglam2023actor}. Learning-progress methods instead prioritize task regions where policy performance improves most rapidly~\citep{portelas2020teacher, zhao2022learning}. Other studies automate curriculum search within parameterized terrain spaces, where tasks are generated by adjusting terrain sampling probabilities, filtering geometric parameters, discretizing stepping-stone configurations, or updating curriculum ranges in a learned latent space~\citep{li2026scaling, lee2020learning, ALLSTEPs, kim2025high}. While these methods automate task selection, they still rely on manually parameterized terrain or task spaces, limiting their applicability to curriculum generation on unstructured maps.

\section{Method}
\begin{figure*}[t]
    \centering
    \includegraphics[width=\textwidth]{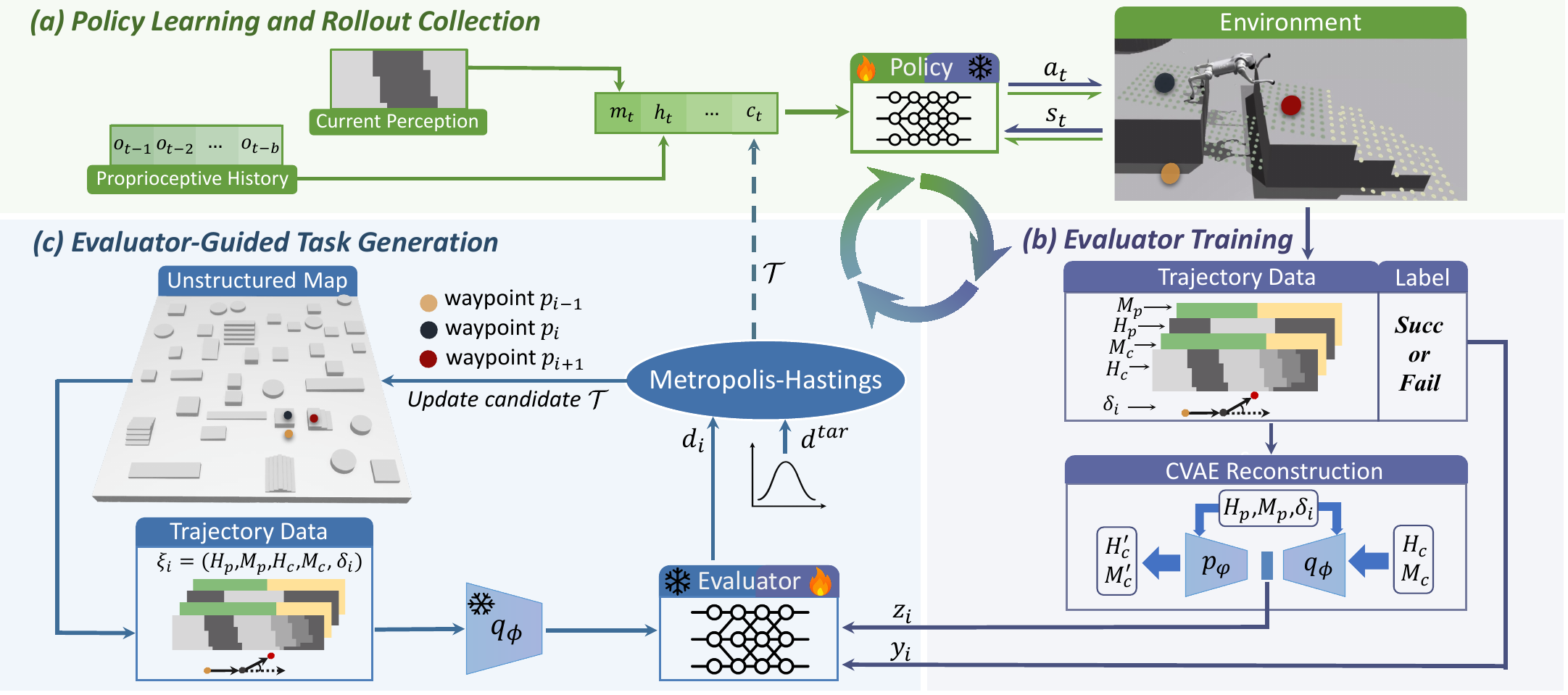}
    \caption{
    \textbf{Framework overview.}
(a) The policy learns to follow waypoint trajectories, where a trajectory \(\mathcal{T}\) is converted into velocity commands \(c_t\) for policy training.
(b) The current policy is rolled out on randomly sampled trajectories to collect transition descriptors \(\xi_i\) and success/failure labels \(y_i\). The CVAE encoder \(q_\phi\) maps each descriptor to a latent code \(z_i\), and the evaluator predicts its policy-conditioned difficulty \(d_i\).
(c) Guided by the updated evaluator, the sampler uses MH-MCMC to generate capability-matched trajectories by sampling for transitions whose predicted difficulty \(d_i\) matches the target difficulty \(d^{\mathrm{tar}}\). The generated trajectories are used for the next policy update, closing the curriculum loop. Fire and snowflake icons indicate training and inference modes.
}
    \label{fig:overview}
    \vspace{-0.8em}
\end{figure*}

As shown in Fig.~\ref{fig:overview}, our framework forms a closed-loop curriculum over trajectory-following tasks. Each curriculum iteration, indexed by \(k\), consists of three coupled processes: (a) policy learning and rollout collection, (b) difficulty evaluator training, and (c) evaluator-guided task generation. Given the current policy \(\pi_{\theta_k}\), rollouts on sampled tasks provide transition-level outcomes for training the evaluator \(\mathcal{D}_k\). The updated evaluator guides the sampler \(\mathcal{G}_k\) to generate the next task set \(\mathcal{S}_{k+1}\), which is used for the subsequent policy update.

\subsection{Problem Formulation}
We formulate our method as an automatic curriculum learning framework that constructs training task sets for legged locomotion on unstructured terrain. Given a terrain map \(\mathcal{M}\), at curriculum iteration \(k\), the framework provides a set of trajectory-following tasks \(\mathcal{S}_k=\{\mathcal{T}_j\}\) for policy training. Each task \(\mathcal{T}=\{p_0,p_1,\ldots,p_N\}\) specifies one rollout episode, where \(p_i\in\mathbb{R}^2\) is a waypoint on \(\mathcal{M}\). The robot starts from \(p_0\) and sequentially tracks \(p_1,\ldots,p_N\). Consecutive waypoints define sub-trajectories \(\tau_i=(p_i,p_{i+1})\), for \(i=0,\ldots,N-1\), where executing \(\tau_i\) means reaching \(p_{i+1}\) from \(p_i\). The policy is trained with DRL: at time \(t\), the actor receives an observation \(s_t\), samples an action \(a_t\sim\pi_\theta(\cdot\mid s_t)\), and obtains reward \(r_t\) after the action is applied. We use the same waypoint-tracking objective and the same asymmetric actor-critic design as the teacher-stage setup of Extreme Parkour~\citep{cheng2024extreme}, details of observations, rewards, and policy hyperparameters are provided in Appendix.

To evaluate and generate curriculum tasks, we represent each sub-trajectory by a transition context descriptor \(\xi_i=(H_p,M_p,H_c,M_c,\delta_i)\). Here \((H_p,M_p)\) denotes the height patch and valid-length mask of the previous sub-trajectory \(\tau_{i-1}=(p_{i-1},p_i)\), and \((H_c,M_c)\) denotes the corresponding pair of the current sub-trajectory \(\tau_i=(p_i,p_{i+1})\), and \(\delta_i\) is the relative yaw angle from \(\tau_{i-1}\) to \(\tau_i\). The height patches satisfy \(H_p,H_c\in\mathbb{R}^{L\times W}\), and the masks satisfy \(M_p,M_c\in\{0,1\}^{L\times W}\). Each patch is cropped along its waypoint direction and start-referenced by subtracting the mean elevation near the beginning of its centerline, while the mask marks the cells covered by the actual transition length. For \(i=0\), the previous context is set to a flat patch, a zero mask, and zero yaw.

\subsection{Difficulty Evaluator Training}
A curriculum update is triggered once the policy reaches an average waypoint-completion threshold \(\kappa_{\mathrm{wp}}\) on the current tasks, or once the update period \(K_{\mathrm{upd}}\) is met. We then evaluate the current policy \(\pi_{\theta_k}\) on feasible trajectories randomly sampled from \(\mathcal{M}\), collecting transition-level rollout outcomes for evaluator training. For each evaluated sub-trajectory \(\tau_i=(p_i,p_{i+1})\), we set \(\ell_i=\|p_{i+1}-p_i\|_2\) and \(T_i=\ell_i/v+\epsilon\), where \(\epsilon\) is a small time tolerance, and define
\begin{equation}
y_i =
\begin{cases}
0, & \text{if } p_{i+1} \text{ is reached within } T_i,\\
1, & \text{if timeout, falling, or another failure termination occurs}.
\end{cases}
\label{eq:yi-definition}
\end{equation}
The label \(y_i\) records the execution outcome of \(\tau_i\) under the current policy, with \(0\) for successful traversal and \(1\) for failure. To balance the training data for \(\mathcal{D}_k\), we store evaluated transitions \(u_i=(\xi_i,y_i)\) in equal-size success and failure queues. Given these labeled transitions,
the evaluator \(\mathcal{D}_k\) learns to estimate the difficulty of a proposed sub-trajectory for the current policy \(\pi_{\theta_k}\). It takes a latent transition code \(z_i\) as input and predicts a score \(d_i=\mathcal{D}_k(z_i)\in[0,1]\), where larger values indicate a higher failure likelihood under \(\pi_{\theta_k}\), and therefore a more difficult transition for the current policy. Instead of predicting this score directly from the high-dimensional transition descriptor \(\xi_i\), we first learn a compact latent representation with a CVAE. Let \(x_i=(H_c,M_c)\) denote the current sub-trajectory representation and \(c_i=(H_p,M_p,\delta_i)\) denote its transition context. The CVAE consists of an encoder \(q_\phi(z_i\mid x_i,c_i)\) and a decoder \(p_\varphi(x_i\mid z_i,c_i)\), with prior \(p(z_i)=\mathcal{N}(0,I)\). We denote the reconstruction produced by the decoder as \(x'_i=(H'_c,M'_c)\). The CVAE is trained by minimizing
\begin{equation}
\mathcal{L}_{\mathrm{cvae}}
=
\lambda_H
\|M_c\odot(H_c-H'_c)\|_2^2
+
\lambda_M
\mathrm{BCE}(M_c,M'_c)
+
\lambda_{\mathrm{KL}}
D_{\mathrm{KL}}
\left(
q_\phi(z_i\mid x_i,c_i)
\,\|\,p(z_i)
\right),
\label{eq:cvae-loss}
\end{equation}
where \(\mathrm{BCE}(\cdot,\cdot)\) denotes point-wise binary cross-entropy. The first term reconstructs height only over the valid region, the second term reconstructs the valid-region mask, and the KL term regularizes the latent space. After CVAE training, the encoder provides \(z_i\) from \((x_i,c_i)\), and the evaluator \(\mathcal{D}_k\) is trained on these latent codes with binary cross-entropy:
\begin{equation}
\mathcal{L}_{\mathrm{diff}}
=
-\frac{1}{B}
\sum_{i=1}^{B}
\left[
y_i \log d_i
+
(1-y_i)\log(1-d_i)
\right],
\qquad
d_i=\mathcal{D}_k(z_i).
\end{equation}
Since \(y_i=1\) corresponds to failed traversal under \(\pi_{\theta_k}\), \(d_i\) measures the difficulty of \(\tau_i\) for the current policy rather than intrinsic terrain difficulty.

\subsection{Evaluator-guided Task Generation}
After updating the evaluator \(\mathcal{D}_k\), the sampler \(\mathcal{G}_k\) generates the next training trajectories by searching over waypoint locations on \(\mathcal{M}\). We formulate task generation as energy minimization problem. For a candidate trajectory \(\mathcal{T}=\{p_0,p_1,\ldots,p_N\}\), the energy is
\begin{equation}
E(\mathcal{T})
=
E_{\mathrm{hard}}(\mathcal{T})
+
\underbrace{
\sum_{i=0}^{N-1}
\left|
d_i-d^{\mathrm{tar}}
\right|
}_{E_{\mathrm{soft}}(\mathcal{T})},
\qquad
d_i=\mathcal{D}_k(z_i),\quad
z_i=\mathrm{Enc}_\phi(x_i,c_i).
\label{eq:sampler-energy}
\end{equation}
Here \(E_{\mathrm{hard}}\) imposes large penalties on infeasible proposals, including trajectories that leave the map boundary, exceed the valid transition length, or place waypoints in invalid or non-traversable regions. \(E_{\mathrm{soft}}\) is the curriculum objective that encourages all sub-trajectories in \(\mathcal{T}\) to match the assigned target difficulty \(d^{\mathrm{tar}}\). For each candidate trajectory, we sample \(d^{\mathrm{tar}}\) from a truncated Gaussian distribution over \([\alpha,\beta]\), centered at moderate difficulty, to emphasize challenging but learnable tasks. In practice, batch sampling draws multiple targets in parallel to improve coverage within this difficulty band. We optimize \(E(\mathcal{T})\) with MH-MCMC for \(J\) iterations. At iteration \(j\), the sampler proposes a new trajectory \(\mathcal{T}'\sim Q(\mathcal{T}'\mid\mathcal{T})\) by perturbing waypoint locations with decaying Gaussian noise, recomputes the affected transition descriptors, and evaluates \(E(\mathcal{T}')\). The proposal is accepted with probability
\begin{equation}
A(\mathcal{T}\rightarrow\mathcal{T}')
=
\min
\left(
1,
\frac{
\exp\left(-E(\mathcal{T}')/T_j\right)
Q(\mathcal{T}\mid\mathcal{T}')
}{
\exp\left(-E(\mathcal{T})/T_j\right)
Q(\mathcal{T}'\mid\mathcal{T})
}
\right),
\label{eq:mh-accept}
\end{equation}
where \(Q(\mathcal{T}'\mid\mathcal{T})\) is the waypoint proposal distribution and \(T_j\) is the sampling temperature. Since our waypoint proposal uses symmetric Gaussian perturbations, we have
\(Q(\mathcal{T}'\mid\mathcal{T})=Q(\mathcal{T}\mid\mathcal{T}')\), so the proposal terms cancel in the Metropolis-Hastings ratio. In our implementation, both the proposal noise scale and \(T_j\) decay over MH-MCMC iterations. Details of \(E_{\mathrm{hard}}\), the proposal distribution, and the decay schedules are provided in Appendix. The target interval \([\alpha,\beta]\) remains fixed across curriculum iterations, but it is interpreted through the current evaluator. As \(\mathcal{D}_k\) is updated from new rollouts, terrain that was difficult for an earlier policy can receive a lower score for the improved policy. Thus, minimizing \(E_{\mathrm{soft}}\) with the same \(d^{\mathrm{tar}}\) drives the sampler toward transitions that remain challenging for the current policy. The final accepted trajectories become the task set for the next policy update, closing the curriculum loop.

\section{Experiment}
\subsection{Experimental Setup}
We design our experiments to answer three questions:
\textbf{Q1:} Can our \ourname{} framework provide an effective curriculum that guides progressive policy learning across a unstructured terrain map? \textbf{Q2:} How does \ourname{} compare with handcrafted curriculum learning method? \textbf{Q3:} Is the CVAE-based transition encoding necessary for reliable difficulty estimation?

\begin{figure*}[t]
    \centering
    \includegraphics[width=\textwidth]{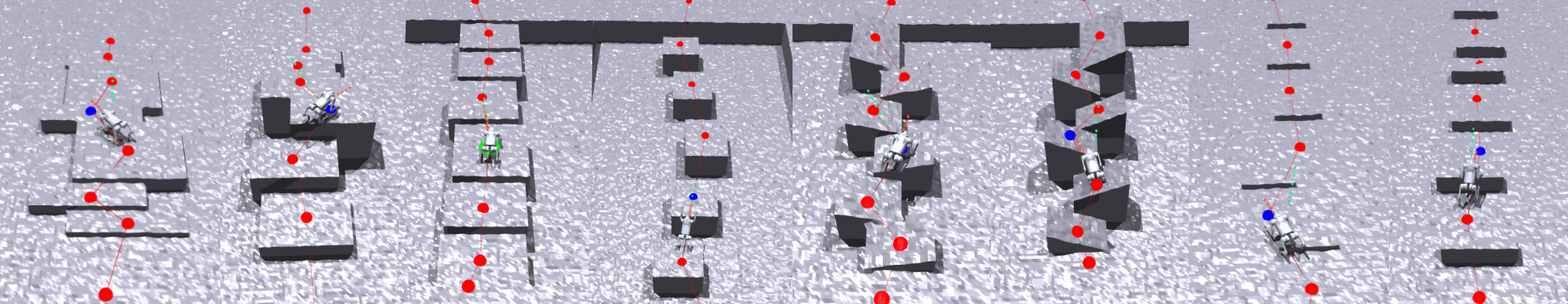}
    \caption{
    Obstacle families in \textit{Map\_HCL}. 
    From left to right: steps, gaps, parkour, and hurdles, each shown at two difficulty levels. 
    Red lines and points denote the predefined waypoint trajectories used in the handcrafted curriculum.
    }
    \label{fig:map_hcl_vis}
    \vspace{-0.3em}
\end{figure*}

\textbf{Terrain Maps.}
We conduct experiments on two terrain maps. \textit{Map\_A} is an unstructured training map with circular platforms, rectangular blocks, and obstacle heights increasing from \(0.05\,\mathrm{m}\) to \(0.6\,\mathrm{m}\) along the map. \textit{Map\_HCL} is a handcrafted curriculum map procedurally instantiated from the Extreme Parkour codebase~\citep{cheng2024extreme}. It contains flat regions and four obstacle families, including steps, gaps, parkour, and hurdles, as visualized in Fig.~\ref{fig:map_hcl_vis}. Each family is organized into \(10\) manually specified difficulty levels, with \(8\) handcrafted sub-terrain instances per level and one predefined \(8\)-waypoint training trajectory per instance.

\textbf{Baselines.}
We compare \ours{} with two baselines. All three methods use the same policy implementation based on the teacher-stage setup of Extreme Parkour~\citep{cheng2024extreme}. The only difference is the source of training tasks. \ours{} samples trajectories adaptively according to the evaluated capability of the current policy. \randst{} trains on feasible trajectories randomly sampled from the given map. \hcle{} follows the original handcrafted curriculum setting of Extreme Parkour, where the policy is trained on predefined waypoint trajectories following a manually designed difficulty progression. All policies are trained and evaluated in parallel simulation using Isaac Gym~\citep{makoviychuk2021isaac}. During evaluation, success is measured at the trajectory level. A rollout is successful only if the robot reaches all waypoints, and it is counted as failed if any sub-trajectory times out, falls, or triggers another failure termination as defined in Eq.~\ref{eq:yi-definition}. Details of all model architectures, hyperparameters, evaluator training, and experimental settings are provided in Appendix.

\textbf{Real-World Experiments.}
For real-world deployment, we follow the sim-to-real pipeline of Extreme Parkour~\citep{cheng2024extreme}. The stage-one policy trained with height-map observations serves as the teacher, and we distill it into a student policy using depth images as exteroceptive input. We deploy the student policy on a Unitree Go2 equipped with an Intel RealSense D435i. This setup isolates our curriculum generation method while using a standard deployment pipeline. Details of stage-two training, deployment, and additional real-world experiments are provided in Appendix.

\begin{figure*}[t]
    \centering
    \includegraphics[width=\textwidth]{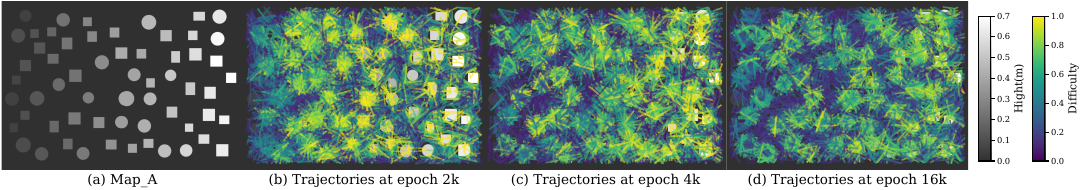}
    \caption{
    Sampler-generated curriculum by our \ourname{} on \textit{Map\_A}.
    (a) The terrain height field.
    (b)--(d) Trajectories sampled by our method at different policy-training epochs.
    }
    \label{fig:mapa_sampler}
    \vspace{-0.8em}
\end{figure*}

\subsection{Results}
\textbf{A1: \ourname{} generates effective curricula that progressively expands policy capability}. 
We train \ours{} and \randst{} on \textit{Map\_A} for \(16\mathrm{k}\) policy epochs. Fig.~\ref{fig:mapa_sampler} visualizes the height map and the trajectories sampled by our method at different epochs. Trajectories are colored by the evaluator-predicted difficulty score, where darker colors indicate easier trajectories and brighter colors indicate harder trajectories. Since the sampler always targets the full difficulty range, the important signal is the spatial location of each difficulty level. At early training, sampled trajectories are limited to flat regions and lower platforms, while the highest platforms are avoided. As the policy improves, the sampled distribution moves toward higher platforms: easy trajectories occupy traversable flat regions, medium trajectories cover platform surfaces, and hard trajectories concentrate near platform boundaries. This shift shows that our method adapts the task distribution to the current policy capability, producing a learnable curriculum that progressively moves toward harder terrain.

\begin{figure*}[t]
    \centering

    \begin{subfigure}[t]{0.48\textwidth}
        \centering
        \includegraphics[width=\linewidth]{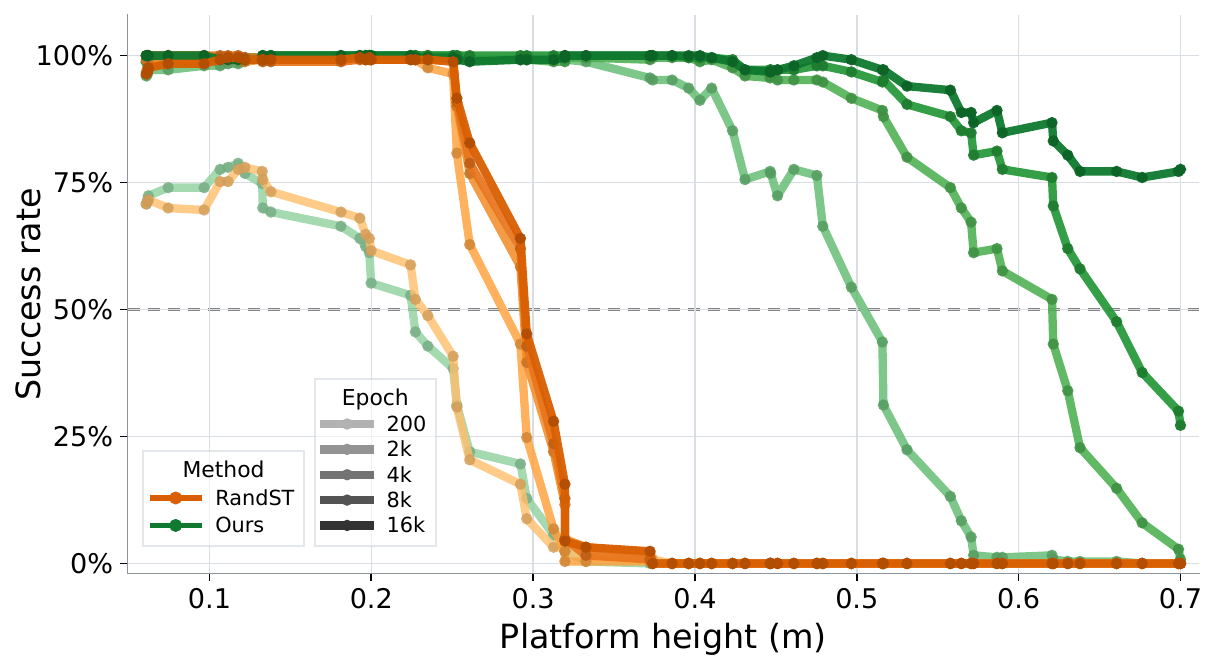}
    \caption{
    Platform-climbing success rate on \textit{Map\_A}. 
    }
        \label{fig:MAP_A}
    \end{subfigure}
    \hfill
    \begin{subfigure}[t]{0.48\textwidth}
        \centering
        \includegraphics[width=\linewidth]{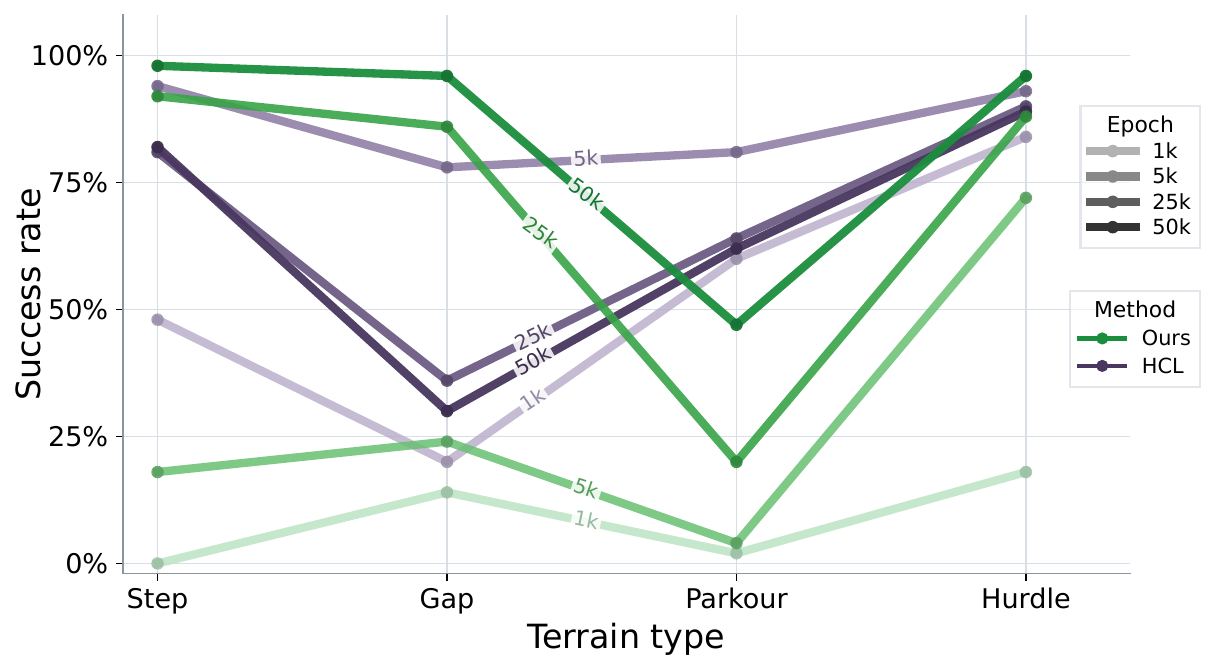}
\caption{Evaluation of policies trained on \textit{Map\_HCL}.}
        \label{fig:MAP_HCL}
    \end{subfigure}

    \caption{(a) Our method \ours{} adapts the training distribution to the evolving policy capability on \textit{Map\_A}. (b) Our method \ours{} achieves comparable results to the handcrafted curriculum baseline on major obstacle families in \textit{Map\_HCL}, without using predefined training trajectories.}
    \label{fig:trajectory_samples}
    \vspace{-0.8em}
\end{figure*}

Fig.~\ref{fig:MAP_A} evaluates policy checkpoints across training epochs on three-waypoint climb-then-descend trajectories, with \(50\) randomly sampled trials per platform. Our \ours{} progressively expands the traversable height range, achieving nonzero success across all heights by \(4\mathrm{k}\), over \(50\%\) success below \(0.65\,\mathrm{m}\) by \(8\mathrm{k}\), and above \(70\%\) success across all heights by \(16\mathrm{k}\). In contrast, \randst{} saturates early and converges to a conservative policy whose reliable climbing capability remains around \(0.3\,\mathrm{m}\). Across all platform heights, our method improves the average success rate from \(39.57\%\) for \randst{} to \(94.54\%\), a \(55.97\%\) gain. These above results show that our method builds a learnable curriculum that progressively improves the policy's traversal capability.

\textbf{A2: Policies trained with \ourname{} generalize better than HCL across most obstacle families.}
We train \hcle{}, \ours{} , and \randst{} on \textit{Map\_HCL} for \(50\mathrm{k}\) policy epochs. \hcle{} follows the handcrafted curriculum protocol of Extreme Parkour~\citep{cheng2024extreme}, using manually specified terrain progressions and predefined \(8\)-waypoint training trajectories. In contrast, \ours{} treats \textit{Map\_HCL} as an ordinary terrain map and adaptively samples training trajectories according to the current policy capability. \randst{} uses randomly sampled feasible trajectories for training. We evaluate policy checkpoints on separate test terrains generated with the same procedural code and difficulty parameters as \textit{Map\_HCL}, but using different random seeds. For each obstacle family, we evaluate the maximum difficulty level on \(8\) test sub-terrain instances, each paired with a corresponding \(8\)-waypoint trajectory, and report the average trajectory-level success rate over \(50\) rollouts.

Fig.~\ref{fig:MAP_HCL} shows that \hcle{} learns quickly from its handcrafted progression but degrades with continued training: from \(5\mathrm{k}\) epochs to the final checkpoint, success changes from \(94\%\!\rightarrow\!82\%\) on step, \(78\%\!\rightarrow\!30\%\) on gap, \(81\%\!\rightarrow\!62\%\) on parkour, and \(93\%\!\rightarrow\!89\%\) on hurdle. This suggests that \hcle{} becomes overly adapted to the predefined trajectories and perceptual patterns. In contrast, our method \ours{} starts with much lower success at \(5\mathrm{k}\) epochs but improves steadily without using handcrafted training trajectories: \(18\%\!\rightarrow\!98\%\) on step, \(24\%\!\rightarrow\!96\%\) on gap, and \(72\%\!\rightarrow\!96\%\) on hurdle. The main limitation is parkour: \(4\%\!\rightarrow\!47\%\), where the trajectory requires a specific human-designed maneuver that is rarely discovered by automatic sampling without additional priors.  Overall, across the four maximum-difficulty obstacle families, our method improves the average success rate from \(65.75\%\) for \hcle{} to \(84.25\%\), an \(18.5\%\) gain. \randst{} achieves \(0\%\) success on all obstacle families and is omitted from the plot for clarity. 

\begin{figure*}[t]
    \centering

    \begin{subfigure}[t]{0.48\textwidth}
        \centering
        \includegraphics[width=\linewidth]{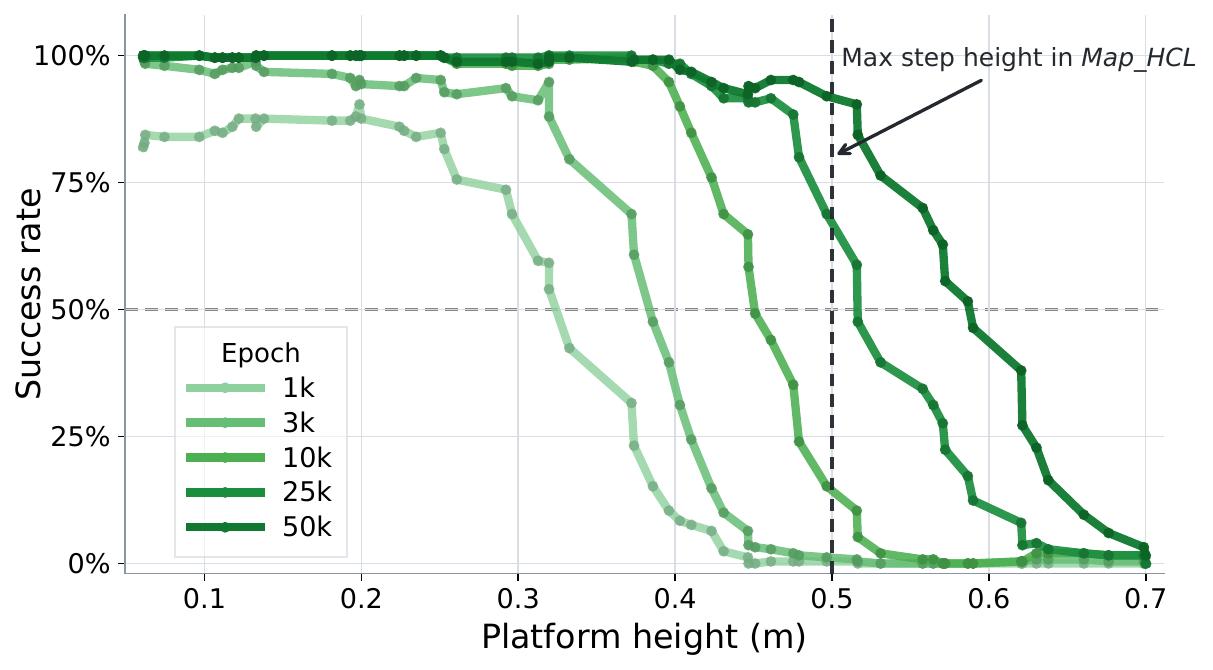}
    \caption{
     Ours, requires no predefined training trajectories.
        }
        \label{fig:robust_ours}
    \end{subfigure}
    \hfill
    \begin{subfigure}[t]{0.48\textwidth}
        \centering
        \includegraphics[width=\linewidth]{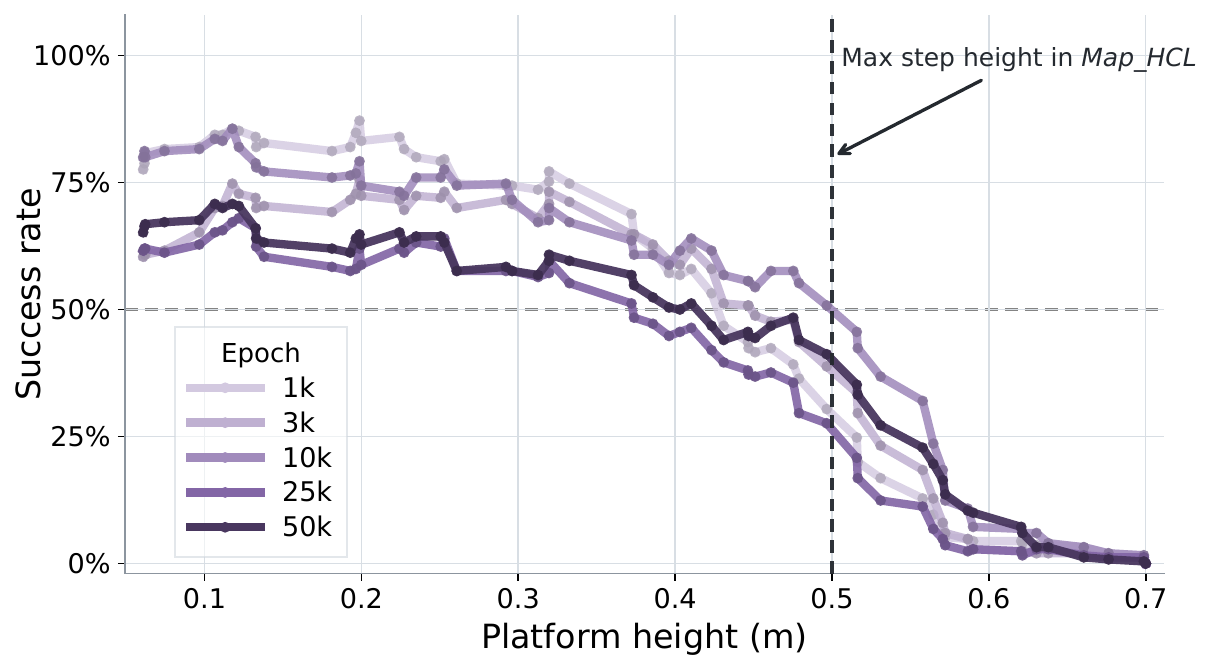}
\caption{HCL, requires predefined training trajectories.}
        \label{fig:robust_hcl}
    \end{subfigure}

\caption{
Platform-climbing evaluation across approach directions.
Each value is the average success rate over diverse approach directions, where our method shows more robust traversal than HCL.
}
    \label{fig:robust}
    \vspace{-0.3em}
\end{figure*}

We further compare \ours{} with \hcle{} on step-task robustness in Fig.~\ref{fig:robust}. In \textit{Map\_HCL}, the highest-difficulty step terrain used for training has a height of \(0.5\,\mathrm{m}\), which marks the maximum step height observed during training. For each policy checkpoint and platform height, we report the average success rate over trajectories approaching the platform from diverse directions. Within this training height range, \ours{} improves the overall success rate from \(58.68\%\) for \hcle{} to \(98.42\%\). Our method shows a capability-frontier expansion strategy, where the policy first improves near its current limit and then shifts the success boundary toward taller platforms. In contrast, \hcle{} shows limited boundary expansion and even degrades at later epochs, suggesting that predefined trajectories provide insufficient variation in approach directions and traversal contexts.

\textbf{A3: CVAE-based transition encoding is critical for reliable difficulty estimation and effective task generation.}
We compare \ours{} with a variant that removes the CVAE representation. 
\begin{wraptable}[7]{r}{0.5\textwidth} 
\centering
\vspace{-10pt}
\caption{Ablation of CVAE-based transition encoding on \textit{Map\_A}.}
\label{tab:cvae_ablation}
\resizebox{\linewidth}{!}{
\begin{tabular}{lccc}
\toprule
Method & Eval. Acc. & Avg. Energy & Success Rate \\
\midrule
\randst{} & -- & -- & \(35.2\%\) \\
\ours{} w/o CVAE & \(67.3\%\) & \(0.013\) & \(47.4\%\) \\
\ours{} w/ CVAE & \(\mathbf{92.5\%}\) & \(0.021\) & \(\mathbf{94.6\%}\) \\
\bottomrule
\end{tabular}
}
\vspace{-0.8em}
\end{wraptable}
In this variant, the evaluator \(\mathcal{D}\) predicts difficulty directly from the raw transition descriptor \(\xi_i=(H_p,M_p,H_c,M_c,\delta_i)\): \((H_p,M_p)\) and \((H_c,M_c)\) are concatenated along the channel dimension, encoded by separate convolutional neural networks (CNNs) \citep{lecun1998gradient}, and fused with \(\delta_i\) for binary difficulty prediction. The sampler and policy-training pipeline remain unchanged. Table~\ref{tab:cvae_ablation} reports evaluator classification accuracy, the average energy of sampled trajectories, and the final trajectory success rate on \(1000\) randomly sampled feasible trajectories in \textit{Map\_A}. Removing the CVAE reduces evaluator accuracy from \(9 2.5\%\) to \(67.3\%\), indicating that direct prediction from raw transition descriptors provides an unreliable difficulty signal. Although the sampler still obtains low energy under this flawed evaluator, the resulting tasks no longer match the capability of the full policy, leading to a success rate drop from 94.6\% to 47.3\%. This degradation suggests that raw height patches form a non-smooth input space for difficulty prediction, where small geometric perturbations can induce large pixel-level changes and make stable geometry-to-difficulty learning difficult from limited rollout data. The CVAE alleviates this issue by encoding transitions into a more compact and smoother latent space, which is critical for training reliable evaluators and generating tasks that match the curriculum.

\section{Conclusion}
We propose \ourname{}, an framework for robot locomotion on unstructured terrain, enabling policies to learn complex traversal skills without handcrafted terrain templates or predefined trajectories. The framework forms a closed loop by learning a difficulty evaluator from current-policy rollouts and using it to guide the sampler in generating new training trajectories for subsequent policy learning. Experiments show that our method generates effective curricula on raw terrain maps, improves robot traversal capability and achieves more robust generalization than HCL.

\section{Limitations}
Although our framework learns robust locomotion behaviors without manually predefined trajectories, highly structured maneuvers may require more directed exploration than generic trajectory sampling provides. Incorporating weak task priors or demonstration-guided proposals could improve sampling efficiency for such behaviors while preserving direct curriculum generation on unstructured maps.

\clearpage

\bibliography{example}  

\end{document}